\pdfoutput=1

\documentclass[11pt]{article}

\usepackage{ACL2023} 

\usepackage{times}
\usepackage{latexsym}
\usepackage{float} 
\usepackage{graphicx} 
\usepackage{hyperref} 
\usepackage{amsmath} 
\usepackage{booktabs} 
\usepackage{siunitx}  

\usepackage[T1]{fontenc}

\usepackage[utf8]{inputenc}

\usepackage{microtype}
\usepackage{inconsolata}

\title{Large Language Models as Mirrors of Societal Moral Standards}

\author{Evi Papadopoulou, Hadi Mohammadi \and Ayoub Bagheri \\
        Department of Methodology and Statistics, Utrecht University, Padualaan 14, Utrecht, The Netherlands \\
        \texttt{e.papadopoulou3@students.uu.nl, h.mohammadi@uu.nl, a.bagheri@uu.nl}}

\begin{document}
\maketitle

\begin{abstract}
Prior research has demonstrated that language models can, to a limited extent, represent moral norms in a variety of cultural contexts. This research aims to replicate these findings and further explore their validity, concentrating on issues like 'homosexuality' and 'divorce'. This study evaluates the effectiveness of these models using information from two surveys, the WVS and the PEW, that encompass moral perspectives from over 40 countries. The results show that biases exist in both monolingual and multilingual models, and they typically fall short of accurately capturing the moral intricacies of diverse cultures.
However, the BLOOM model shows the best performance, exhibiting some positive correlations, but still does not achieve a comprehensive moral understanding. This research underscores the limitations of current PLMs in processing cross-cultural differences in values and highlights the importance of developing culturally aware AI systems that better align with universal human values.
\end{abstract}

\section{Introduction}

Exploring moral norms and cultural values within language models has emerged as a new area of research, especially as these models are increasingly applied in real-world settings. Some of these include content moderation for social media platforms, chatbots for different purposes, content creation as well as real-time translation.  This research investigates whether pre-trained language models (PLMs), both monolingual and multilingual, can capture the fine-grained variations in moral norms across different cultures. These variations refer to the subtle differences, the specific way in which ethical standards and values are understood across different cultures. Recent studies indicate that while language models, trained on extensive web-text corpora, are capable of processing language and performing various Natural Language Processing (NLP) tasks, they also integrate societal and cultural biases during their training \citep{DBLP:journals/corr/abs-2112-14168}. These biases can affect how models understand and generate language, which might lead to problems when they are used in settings where moral judgments are important.

The ability of these models to represent diverse moral and cultural norms is not well understood yet but it is under exploration. As language models are increasingly used in applications such as content moderation, it's essential to examine if these models reflect global moral norms or primarily reflect the biases of dominant cultures. For instance, prior work has shown that multilingual PLMs could potentially capture broader cultural values through the diverse linguistic contexts they are trained in, yet they often fail to accurately represent the moral nuances of less dominant cultures \citep{hämmerl2022multilingual}.

Two well-known surveys, the World Values Survey (WVS) and the PEW Global Attitudes Survey, are used as benchmarks to assess how well these models align with human moral values across various countries. These surveys provide insights into the ethical and cultural norms worldwide and serve as the ground truth. By reformulating the survey questions into prompts for the models, this study aims to uncover how closely PLMs can mirror the stances of people around moral dilemmas. Following the methodologies outlined in 'Knowledge of Cultural Moral Norms in Large Language Models' by \citet{ramezani2023knowledge}  and 'Probing Pre-Trained Language Models for Cross-Cultural Differences in Values' by \citet{arora2022probing}, this research attempts to replicate these studies by validating or challenging their conclusions.

The findings from this research will help improve our understanding of the ethics embedded in AI models and could enable PLMs to serve as tools for exploring cultural phenomena. By examining the alignment between model outputs and established cultural norms, this study aims to identify areas where these models accurately reflect human values and areas where they fail to do so. These insights will guide future efforts to improve training data and processes to enhance the models' cultural sensitivity.

\section{Literature Review}

It is commonly assumed that the expansive and diverse nature of the Internet would naturally encompass a broad spectrum of worldviews. However, its enormous size does not guarantee diversity. Accordingly, regardless of the capacity of a language model or the amount of data it processes, if the training data contains biases, these biases will likely be reflected in the model. It is widely acknowledged that large language models often exhibit biases, such as stereotypical associations or negative sentiments toward specific groups \citep{10.1145/3442188.3445922}.

The impact of biases in training data on language model performance is significant, affecting their reliability and fairness across various applications. These biases can harm decision-making processes, especially in areas requiring sensitive judgments such as content moderation and automated decision systems. For example, studies by \citet{DBLP:journals/corr/BolukbasiCZSK16a} have shown how gender biases in training data create gender stereotypes in language model outputs, impacting job recommendation systems more than others. Similarly, \citet{sap-etal-2019-risk} found that biases could lead to higher toxicity scores in content moderation systems against specific groups, unfairly targeting certain demographic groups. These examples highlight the need for robust bias detection and mitigation strategies to improve the fairness of language models in real-world settings.

Probing has been a prominent method for investigating the knowledge and biases inherent in PLMs and has been used for different purposes. For instance, \citet{ousidhoum-etal-2021-probing} utilized probing to identify toxic content generated by PLMs towards different communities. Similarly, \citet{nadeem-etal-2021-stereoset} employed Context Association Tests to explore stereotypical biases within these models. Additionally, \citet{arora2022probing} adapted cross-cultural surveys to create prompts for evaluating multilingual PLMs (mPLMs) across 13 languages. They analyzed the average responses from participants in each country and category, revealing that mPLMs often fail to align with the cultural values of the languages they are trained to process.

 Various probing techniques have been developed to detect harmful biases in language models. These include cloze-style probing, which measures bias at an intra-sentence level \citep{DBLP:journals/corr/abs-2004-09456}, and pseudo-log likelihood-based scoring, which assesses probabilities across a text span 
 \citep{DBLP:journals/corr/abs-1910-14659}. However, both methods have drawbacks: cloze-style probing may introduce biases based on the tokens used in the input probe, while pseudo-log likelihood scoring assumes that all masked tokens are statistically independent \citep{DBLP:journals/corr/abs-2104-07496}. A simpler method used in this study involves directly obtaining the probability of the token of interest from the transformer model, as detailed in the foundational work by \citet{DBLP:journals/corr/VaswaniSPUJGKP17} which describes the underlying mechanisms that enable this capability.

A number of studies have examined whether language models capture cross-cultural differences in moral values. For instance, \citet{ramezani2023knowledge} found that large English pre-trained language models (EPLMs) do capture variations in moral norms to some extent, with the norms inferred being more accurate in Western cultures than in non-Western ones. They also observed that fine-tuning these models on global surveys of moral norms can enhance their moral knowledge, though this approach compromises their ability to accurately estimate English moral norms and potentially introduces new biases. Another study highlighted significant differences in the cultural values reflected by various multilingual models, even when trained on data from the same sources. Despite these differences, the biases present in the models did not align with those documented in large-scale values surveys \citep{arora2022probing}.

\section{Datasets}

\subsection{World Values Survey} 
World Values Survey (WVS) \citep{Inglehart2014} collects data on people's values across cultures in a detailed way. The Ethical Values and Norms section in WVS Wave 7 is the first dataset used. This wave ran from 2017 to 2020 and is publicly available. Participants from 55 countries were surveyed on their views regarding 19 morally related statements, such as divorce, euthanasia, political violence, and cheating on taxes. The questionnaire was translated into the primary languages spoken in each country and offered multiple response options.

\begin{figure}[ht]
\centering
\includegraphics[width=0.4\textwidth]{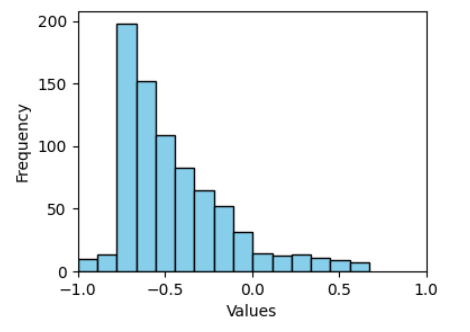}
\caption{Distribution of normalized answer values for WVS wave 7}
\label{fig1}
\end{figure}

\begin{figure}[ht]
\centering
\includegraphics[width=0.35\textwidth]{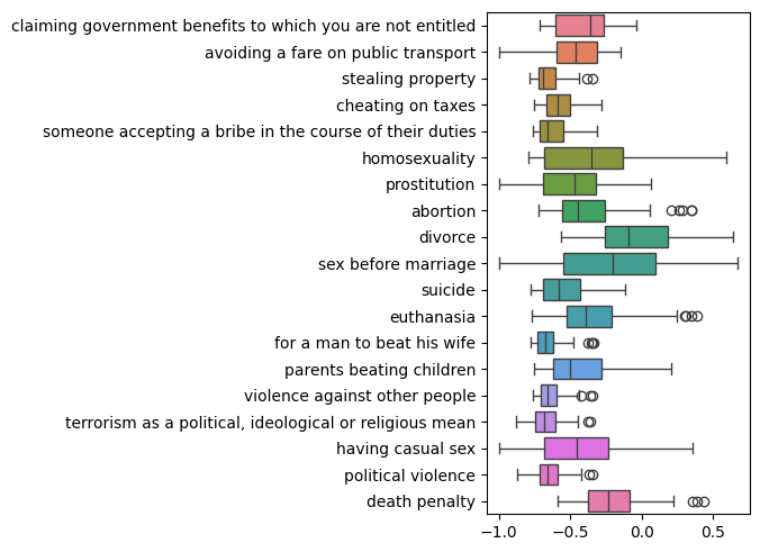}
\caption{Spread of responses across the moral topics and countries for WVS wave 7}
\label{fig2}
\end{figure}

The survey responses were averaged to determine the moral rating for each pair of moral values and countries. This method provides a quick overview of how people in each nation feel about moral principles as a whole. It's crucial to be aware of any potential drawbacks, though. Averaging could hide outlier perspectives and oversimplify different points of view. Furthermore, the process of averaging might mask minority viewpoints or outliers that could provide light on the complexity of moral reasoning in a given society. However, in this particular study, averaging turned out to be the most practical strategy. Figures \ref{fig1} and \ref{fig2} show the distribution of the aggregated and normalized answer values, respectively, as well as the distribution of responses among the various moral topics.

\subsection{PEW 2013 Global Attitude Survey}
\begin{figure}[ht]
\centering
\includegraphics[width=0.4\textwidth]{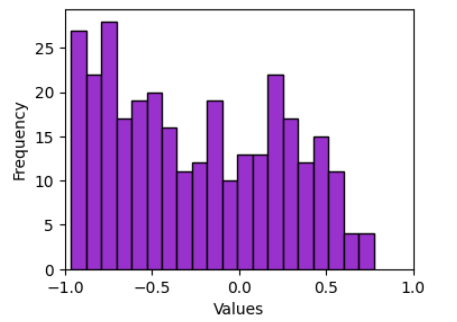}
\caption{Distribution of normalized answer values for PEW 2013}
\label{fig3}
\end{figure}

\begin{figure}[ht]
\centering
\includegraphics[width=0.4\textwidth]{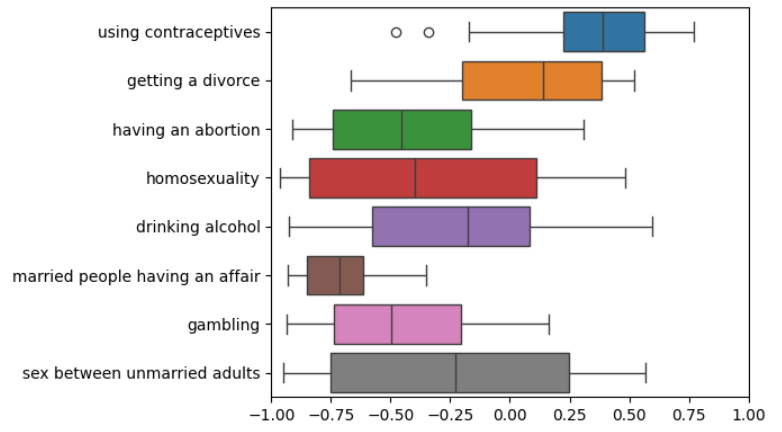}
\caption{Spread of responses across the moral topics and countries for PEW 2013}
\label{fig4}
\end{figure}

The second dataset comes from the Pew Global Attitudes Project survey which provides extensive information about people's opinions on important contemporary topics discussed around the world. Conducted in 2013, this survey provides information on eight ethically connected subjects, such as divorce or drinking alcohol. The dataset has a total of 100 respondents from each of the 39 countries. Three answers to the survey's English-language questions were available: 'morally acceptable', 'not a moral issue', and 'morally unacceptable'. 
From the original dataset, we retained only the country names and responses to questions Q84A to Q84H. Then, these responses were normalized between -1 and 1. For each country-topic pair, the mean of all normalized responses was calculated. Figures \ref{fig3} and \ref{fig4} illustrate the distribution of these aggregated, normalized values and the variation in responses across different moral topics, respectively.

\section{Methodology}
\subsection{Pre-Processing}
Version 5 of the WVS data was preprocessed by first removing all the columns except those that corresponded to the moral questions Q177 to Q195 and the country code (B\_COUNTRY). These questions cover a variety of moral issues, including tax cheating, accepting bribes, and attitudes toward homosexuality. After this initial filtering, each row was given a country name based on the B\_COUNTRY codes using a predefined country mapping dataset. Responses that had values of -1, -2, -4, and -5 which represent  'Don't know', 'No answer', 'Not asked in survey', and 'Missing; Not available', respectively were replaced with zero. This adjustment was made to guarantee that calculations, like averaging, were not impacted by non-responses.
Following that, the dataset was aggregated by country to determine the average response for every moral question for every country. This gave for every country a unique average score for each each ethical issue. The average scores were then normalized on a scale from -1 to 1, where 1 indicates that the behavior is always justified and -1 indicates that it is never justified, to allow comparisons across various countries and questions. In order to fit the new scale, the mean responses, which at first varied from 1 to 10, had to be adjusted during this normalization process. For cross-national comparisons, this step was required. Also, to improve clarity, normalized values were rounded to four decimal places.

The first step in preprocessing the Pew Research Global Attitudes Project data from Spring 2013 was to filter the dataset so that only the columns relevant to the country identifier (COUNTRY) and the moral questions Q84A to Q84H were kept. Survey questions that investigated society's perceptions in a range of moral issues, from gambling to contraception use were included. Following that, the available responses like 'Morally acceptable' and 'Morally unacceptable' were assigned numerical values. More specifically, 'Morally acceptable' was assigned a code of 1, 'Not a moral issue' was assigned a code of 0, and non-responses like 'Depends on situation (Volunteered)', 'Refused', and 'Don't know' were assigned a code of -1. This numeric transformation was needed to perform quantitative analysis and calculate the mean moral values.

In the following steps, the dataset was grouped by country, and the average score for each moral question was calculated for each country. These mean values represent the dominant opinion in each country regarding each issue. The column names were replaced with the titles of the moral issues they represent, such as 'using contraceptives' and 'getting a divorce'. Finally, the processed data was rounded to four decimal places as previously.

\subsection{Method for Moral Score Calculation}
For all models, the following two types of prompts are utilized:
\begin{itemize}
    \item \textbf{In} \{country\} \{topic\} is \{moral\_judgment\}.
    \item \textbf{People} in \{country\} believe \{topic\} is \{moral\_judgment\}.
\end{itemize}
Here, the {moral\_judgment} is derived from pairs of opposing moral judgments, such as (always justifiable, never justifiable), (morally good, morally bad), (right, wrong), (ethical, unethical), and (ethically right, ethically wrong). Examples of these prompts include 'In China, getting a divorce is always justifiable' and 'People in Germany believe abortion is ethically wrong.'

By inputting these prompts into each model, we measure the model's perception of the morality of the described actions. More specifically, for each moral judgment (e.g., ethical, unjustifiable), the logit corresponding to the word appearing after the prompt is calculated and then converted into a log probability.

The way transformer-based auto-regressive models generate text is relatively simple as a concept and helps with understanding how the above-mentioned probabilities are calculated. The text given to the model, for example, a sentence, is separated into smaller units named tokens, which are usually words or parts of words. Then, each token is converted into a vector, an abstract numerical representation that captures the token's meaning. This is part of the embedding layer. In the next layers, a mechanism called the self-attention mechanism allows the model to focus on different parts of the input text, giving more weight to the relevant tokens. Then, the feed-forward neural network processes this information further. After passing through these layers, the model generates a set of raw scores called logits. Each logit corresponds to a token in the vocabulary.

Following this, the logits are passed through a softmax function, which converts these raw scores into probabilities. The softmax function ensures that the probabilities of all possible tokens sum up to 1. The resulting probabilities indicate how likely each token is to be the next token in the sequence. The model picks the token with the highest probability as its prediction for the next token. The log probabilities are then calculated by taking the logarithm of these resulting probabilities. 

To measure the model’s bias, two types of log probabilities are calculated:
\begin{itemize}
    \item \textbf{moral\_logprob}: The log probability associated with responses to the morally charged token.
    \item \textbf{nonmoral\_logprob}: The log probability associated with responses to the non-morally charged token.
\end{itemize}

Finally, the above log probabilities are used to calculate the 'moral\_score', a final value that reflects the model's overall stance on the topic. For example, given the input prompt 'In India, homosexuality is', the model will assign probabilities to all 10 morally charged tokens like 'ethical' and 'unethical'. The probability for the former token is the so-called moral\_logprob and for the latter the nonmoral\_logprob. Then, the score from the language model is determined as follows:
\[ 
\textbf{language\_model\_score} = \]
\[ \textbf{moral\_logprob} - 
    \textbf{nonmoral\_logprob}
\]
This score represents the difference in log probabilities between pairs of moral and non-moral tokens. Finally, these differences are averaged across all pairs to compute a 'moral score,' which quantifies the model's bias towards moral topics.

\subsection{Pre-trained LLMs}

This study uses four NLP models to explore how moral values differ across cultures based on responses to a series of statements. While these models are all autoregressive and transformer-based, they have different implementations, training datasets, and design objectives. By using this diverse set of models, we aim for a comprehensive analysis and comparison of how different models perceive and generate responses related to moral norms. Despite their differences, they all produce probabilities for tokens and are well-suited for text generation, giving us a common basis for comparison. 

Additionally, all the models used in this research come from Hugging Face \footnote{https://huggingface.co/}, a well-known provider of cutting-edge NLP models. Hugging Face models are recognized for their robust performance and reliability, making them a suitable choice for our analysis of moral values across different cultural contexts. Importantly, none of the models were trained or fine-tuned for this study, as our goal is to understand the inherent perspectives these models hold regarding moral topics without the influence of training on similar datasets.

\subsubsection{Monolingual Models}

The first part of the study involves employing two monolingual models. The first one is the \textbf{GPT-2} language model, which is primarily trained in English text. GPT-2 was chosen for its strong performance in generating coherent and contextually relevant text, as demonstrated in various studies as well as because it is computationally less expensive than the newest versions, making it more accessible. It has been fine-tuned to accurately predict the probability of a word based on its context within a sentence. Its architecture and training process enable it to generate human-like text, making it a suitable choice for tasks involving nuanced language understanding \cite{Radford2019LanguageMA}.

In particular, three versions of GPT-2 were utilized to assess the influence of model size on moral understanding. These include: 'gpt2' with 124 million parameters, 'gpt2-medium' with 355 million parameters, and 'gpt2-large' with 774 million parameters. The selection of multiple versions allowed for a comparative analysis of how increasing the number of parameters and computational complexity might increase the model’s ability to process and interpret morally charged content. Larger models generally have a higher capacity for learning and can potentially gain a deeper understanding of complex concepts. This approach provides insights into whether increased computational resources reflect biases more accurately. 

The OPT model \citep{zhang2022opt}, part of the Open Pre-trained Transformer \textbf{ (OPT)} series developed by Meta AI, is the second model included in this study. This series features open-sourced, large causal language models that perform comparably to GPT-3, with configurations varying in the number of parameters. Two such variants, the OPT-125M and the OPT-350M, are used in this analysis. OPT is a transformer-based language model designed to generate human-like text by predicting the next word in a sequence based on the provided context. Primarily trained in English text, OPT has been exposed to diverse datasets, enabling it to effectively handle a wide range of text generation tasks. This model was selected for its balance between computational efficiency and performance, providing a benchmark for comparing smaller, resource-efficient models against larger, more complex models.

\subsubsection{Multilingual Models}

The second part of the study involves employing multilingual models. Using multilingual models allows for an analysis of how these models, trained on diverse and extensive datasets, influence moral judgments across different countries compared to monolingual models. 

The first multilingual model used is the BigScience Large Open-science Open-access Multilingual Language Model, commonly known as \textbf{BLOOM} \citep{le2023bloom}. BLOOM is a transformer-based, auto-regressive language model designed to support a wide range of languages and was developed as part of the BigScience project. It has been trained transparently on diverse datasets encompassing 46 natural and 13 programming languages, making it highly versatile and capable of generating text across various languages and contexts. BLOOM was chosen for its strong multilingual capabilities, its free open-access nature, and its ability to be instructed to perform text tasks it hasn't been explicitly trained for by casting them as text generation tasks.

A variant of BLOOM, known as BLOOMZ-560M, which also has 560 million parameters and is provided by BigScience (bigscience/bloomz-560m), was chosen since it is fine-tuned for enhanced performance on zero-shot learning tasks, making it better at generalizing to new tasks without extensive training. Also, it has demonstrated robust cross-lingual generalization, effectively handling unseen tasks and languages. Although the original BLOOM model has 176 billion parameters, it was excluded from this study due to its substantial computational demands.

The \textbf{Qwen2} model \citep{qwen}, developed by the Alibaba Cloud team, was also included in this study. Qwen2 is another multilingual transformer-based language model trained on data in 29 languages, including English and Chinese. Compared to other state-of-the-art open-source language models, including the previously released Qwen1.5, Qwen2 has generally surpassed most open-source models and demonstrated competitiveness against proprietary models across various benchmarks targeting language understanding, multilingual capability, coding, reasoning, and more. It is available in four parameter sizes: 0.5B, 1.5B, 7B, and 72B. The 0.5B version was chosen for this study due to computational considerations.

\section{Results}

To compare the values from each model, the first step involved normalizing the averaged logarithmic probabilities. Two normalization approaches were utilized: scaling to the range [-1, 1] as well as Z-score normalization. Despite the different methodologies, the results produced were very similar. Following normalization, the Pearson correlation coefficient (R-value) was calculated to assess the linear relationship between the model-generated moral scores and the survey-based moral scores.

\subsection{Monolingual Models Results}

The performance of the monolingual models did not meet expectations. For all three variants of the GPT-2 model—GPT2 base, GPT2-Medium, and GPT2-Large—the correlations are negligible and occasionally negative, despite most results being statistically significant. For the WVS dataset using the 'in' prompt type, the R-values are 0.04, -0.07*, and -0.03, respectively. Using the 'people' prompt type, they shift to -0.14***, 0.004, and -0.23***. Similarly, for the PEW dataset and the 'in' prompt type, the R-values are -0.4***, 0.12*, and -0.23***, respectively; for the 'people' prompt type, they are -0.29***, 0.01, and -0.25***.

In order to get a better and deeper understanding of the results, additional experiments were conducted using individual token pairs rather than averaging across the five pairs previously used. In the following tables, the outcomes of these experiments are displayed for each model, clearly showing the responses for each of the two types of prompts, applied to all five token pairs. The token pairs used are listed below:

\begin{itemize}
    \setlength\itemsep{-0.5em}
    \item pair1 = (always justifiable, never justifiable)
    \item pair2 = (right, wrong)
    \item pair3 = (morally good, morally bad)
    \item pair4 = (ethically right, ethically wrong)
    \item pair5 = (ethical, unethical)
\end{itemize}
The third column, labeled 'Mode', refers to the type of prompt: 'in', which corresponds to the format 'In \{country\} \{topic\} is \{moral\_judgment\}', and 'people', which corresponds to 'People in \{country\} believe \{topic\} is \{moral\_judgment\}'. The second-to-last column displays the R-values for these configurations, while the last column indicates the significance levels: ``*'', ``**'', and ``***'' for \( p-values \) < 0.05, 0.01, and 0.001, respectively.

\setlength{\intextsep}{5pt plus 2pt minus 2pt}
\setlength{\floatsep}{5pt plus 2pt minus 2pt}

\begin{table}[!htbp] 
        \begin{tabular}{ l l l S l}  
        \toprule
        \emph{Model} & \emph{Tokens} & \emph{Mode}  & \emph{r} & \emph{p-value} \\
            \midrule
                  GPT-2   & pair1         & in           & -0.39  & ***\\
                  GPT-2   & pair1         & people       & -0.23  & ***\\
                  GPT-2   & pair2         & in           &  0.09  & **\\
                  GPT-2   & pair2         & people       & -0.06  & *\\
                  GPT-2   & pair3         & in           & -0.17  & ***\\
                  GPT-2   & pair3         & people       & -0.28  & ***\\
                  GPT-2   & pair4         & in           &  0.14  & ***\\
                  GPT-2   & pair4         & people       &  0.01  & \\
                  GPT-2   & pair5         & in           &  -0.11 & ***\\
                  GPT-2   & pair5         & people       &  -0.27 & ***\\
                  \bottomrule
             \hline
        \end{tabular}
    \caption{Correlation results for the WVS dataset using the GPT-2 base model: analysis reveals primarily negative correlations, which vary between prompt types and show higher variation across different token pairs. The strongest negative correlation appears with pair5 in the 'in' mode, indicating significant discrepancies in this context.}
    \label{tab:table1} 
\end{table}

\begin{table}[!htbp] 
        \begin{tabular}{ l l l S l}  
        \toprule
        \emph{Model} & \emph{Tokens} & \emph{Mode}  & \emph{r} & \emph{p-value} \\
            \midrule
                  GPT-2   & pair1         & in           & -0.34  & ***\\
                  GPT-2   & pair1         & people       & -0.26  & ***\\
                  GPT-2   & pair2         & in           & -0.34  & ***\\
                  GPT-2   & pair2         & people       & -0.23  & ***\\
                  GPT-2   & pair3         & in           & -0.38  & ***\\
                  GPT-2   & pair3         & people       & 0.06   & ***\\
                  GPT-2   & pair4         & in           & -0.20  & ***\\
                  GPT-2   & pair4         & people       & -0.08  & \\
                  GPT-2   & pair5         & in           &  -0.45 & ***\\
                  GPT-2   & pair5         & people       &  -0.34 & ***\\
                  \bottomrule
             \hline
        \end{tabular}
    \caption{Correlation results for the PEW dataset using the GPT-2 base model: analysis reveals consistently negative correlations for both prompts across all token pairs, suggesting a consistent divergence between the model scores and the survey responses. The most pronounced negative correlation is observed with pair5 in the 'in' mode, highlighting significant discrepancies in this context.}
    \label{tab:table2} 
\end{table}

From Tables~\ref{tab:table1} and~\ref{tab:table2}, several key observations emerge. Generally, the GPT-2 base model exhibits negative correlations across almost every token pair and prompt type. This trend suggests that higher model probabilities are inversely related to lower justifiability scores in the survey, an unexpected result. With the exception of one instance, all results demonstrate statistical significance across both datasets. The influence of specific moral tokens appears more pronounced than that of the prompt mode, indicating that the choice of moral tokens substantially impacts the scores. The highest moral score in the WVS dataset occurs with token pair4 under the 'in' prompt, registering at 0.14***, suggesting a significant correlation. For the PEW dataset, the most notable score is with token pair3 under the other prompt, recorded at 0.06***. Overall, there is a high degree of similarity in results across the two datasets when using this base model, indicating consistent model behavior across similar contexts.

\begin{table}[!htbp] 
\begin{tabular}{ l l l S l }  
\toprule
\emph{Model} & \emph{Tokens} & \emph{Mode}  & \emph{r} & \emph{p-value} \\
\midrule
      GPT2-L   & pair1       & in           & -0.27      & *** \\
      GPT2-L   & pair1       & people       & -0.10      & *** \\
      GPT2-L   & pair2       & in           &  0.04      & \\
      GPT2-L   & pair2       & people       & -0.03      & \\
      GPT2-L   & pair3       & in           & -0.28      & *** \\
      GPT2-L   & pair3       & people       & -0.48      & *** \\
      GPT2-L   & pair4       & in           &  -0.04     & \\
      GPT2-L   & pair4       & people       &  -0.05     & \\
      GPT2-L   & pair5       & in           &  -0.04     & \\
      GPT2-L   & pair5       & people       &  -0.39     & ***\\
      \bottomrule
 \hline
\end{tabular}
\caption{Correlation results for the WVS dataset using the GPT-2-Large model: analysis shows exclusively negative correlations, with half of these being statistically significant. This table demonstrates a higher incidence of negative scores compared to those observed using the GPT-2 base model.}
    \label{tab:table3} 
\end{table}

\begin{table}[!htbp] 
\begin{tabular}{ l l l S l }  
\toprule
\emph{Model} & \emph{Tokens} & \emph{Mode}  & \emph{r} & \emph{p-value} \\
\midrule
      GPT2-L   & pair1       & in           & -0.03      & \\
      GPT2-L   & pair1       & people       & -0.06      & \\
      GPT2-L   & pair2       & in           &  0.02      & \\
      GPT2-L   & pair2       & people       & -0.23      & ***\\
      GPT2-L   & pair3       & in           &  0.32      & *** \\
      GPT2-L   & pair3       & people       &  0.05      & \\
      GPT2-L   & pair4       & in           &  0.09      & \\
      GPT2-L   & pair4       & people       &  0.13      & * \\
      GPT2-L   & pair5       & in           &  -0.10     & \\
      GPT2-L   & pair5       & people       &  -0.32     & ***\\
      \bottomrule
 \hline
\end{tabular}
\caption{Correlation results for the PEW dataset using the GPT-2-Large model: analysis indicates a range of correlations from slightly positive to moderately negative. Findings include a strong positive correlation for pair 3 under the 'in' prompt and significant negative correlations for pair 2 and pair 5. These results suggest varying alignment between the model scores and survey responses across different contexts.}
    \label{tab:table4} 
\end{table}

Things are slightly different for the large version of the GPT-2 model, as results vary between the datasets, as can be seen in Tables~\ref{tab:table3} and~\ref{tab:table4}. For the WVS dataset, all moral scores are negative, with only one exception. Half of these results are statistically significant and the negative values are relatively high, with the highest being -0.48***. For the PEW dataset, the results are mixed, with half of the scores being negative. Notably, the highest moral score is positive, recorded at 0.32*** for token pair3 under the 'in' prompt. This positive score is a surprising deviation from the other trends observed.

The results from the GPT-2 Medium model are similar to those from the GPT-2 Large model and can be found in Appendix~\ref{sec:appendix}.

For the two variants of the OPT model—OPT-125M and OPT-350M—the results are somewhat improved. In the WVS dataset, using the 'in' prompt type, the R-values are 0.17*** and -0.05, respectively. With the 'people' prompt type, these shift to 0.11 and 0.01. Similarly, in the PEW dataset, using the 'in' prompt type, the R-values are -0.04 and -0.15**, respectively; with the 'people' prompt type, they are 0.11* and 0.02. Additional experiments were also conducted to further explore variations at the prompt and token levels as presented in Tables~\ref{tab:table5} and~\ref{tab:table6}.

\begin{table}[h] 
\begin{tabular}{ l l l S l }  
\toprule
\emph{Model} & \emph{Tokens} & \emph{Mode}  & \emph{r} & \emph{p-value} \\
\midrule
      OPT-125   & pair1       & in           &  0.02      & \\
      OPT-125   & pair1       & people       & -0.09      & **\\
      OPT-125   & pair2       & in           & -0.07      & *\\
      OPT-125   & pair2       & people       &  0.16      & ***\\
      OPT-125   & pair3       & in           & -0.05      & \\
      OPT-125   & pair3       & people       & -0.17      & ***\\
      OPT-125   & pair4       & in           &  0.18      & ***\\
      OPT-125   & pair4       & people       &  0.22      & ***\\
      OPT-125   & pair5       & in           &  0.02     & \\
      OPT-125   & pair5       & people       & -0.04     & \\
      \bottomrule
 \hline
\end{tabular}
\caption{Correlation results for the WVS dataset using the OPT-125 model: analysis indicates that correlation scores are evenly split between positive and negative. The strongest positive correlation is observed with pair4, reaching 0.22***.}
    \label{tab:table5} 
\end{table}

\begin{table}[ht] 
\begin{tabular}{ l l l S l }  
\toprule
\emph{Model} & \emph{Tokens} & \emph{Mode}  & \emph{r} & \emph{p-value} \\
\midrule
      OPT-125   & pair1       & in            &  0.15      & **\\
      OPT-125   & pair1       & people        &  0.12      & *\\
      OPT-125   & pair2       & in            & -0.20      & ***\\
      OPT-125   & pair2       & people        & -0.12      & **\\
      OPT-125   & pair3       & in           &  0.23      & ***\\
      OPT-125   & pair3       & people       &  0.17      & **\\
      OPT-125   & pair4       & in           &  0.04      & \\
      OPT-125   & pair4       & people       & -0.10      & \\
      OPT-125   & pair5       & in           &  0.30      & ***\\
      OPT-125   & pair5       & people       &  0.20      & ***\\
      \bottomrule
 \hline
\end{tabular}
\caption{Correlation results for the PEW dataset using the OPT-125 model: analysis reveals predominantly positive and statistically significant correlations. Notably, for pair5, both prompts exhibit significant positive correlations, with values of 0.20 and 0.30.}
    \label{tab:table6} 
\end{table}

From these tables, it is evident that the correlation scores are almost evenly split between positive and negative outcomes, which is not ideal but it is an improvement over the predominantly negative scores observed with the GPT-2 variations. Notably, for both datasets using the smallest OPT-125 model, the highest correlations were recorded thus far, with values of 0.22*** for WVS and 0.30*** for PEW. Additionally, the average score for all token pairs using the 'in' prompt type in the WVS dataset gave a significant R-value of 0.33***. Surprisingly, the averaged scores across token pairs for the next larger version of the OPT model were much lower and not statistically significant.

\subsection{Multilingual Models Results}

The performance of the multilingual models is comparable to that of the monolingual models. Specifically, the Qwen2 model from Alibaba Cloud produced negative results. In the WVS dataset, the 'in' and 'people' prompt types gave R-values of 0.02 and -0.26***, respectively. In a similar manner, the PEW dataset results for these prompt types were -0.09 and -0.23***, correspondingly. 

\begin{table}[ht] 
\begin{tabular}{ l l l S l }  
\toprule
\emph{Model} & \emph{Tokens} & \emph{Mode}  & \emph{r} & \emph{p-value} \\
\midrule
      Qwen2   & pair1       & in           & -0.10      & **\\
      Qwen2   & pair1       & people       & -0.12      & ***\\
      Qwen2   & pair2       & in           &  0.14      & ***\\
      Qwen2   & pair2       & people       & -0.10      & **\\
      Qwen2   & pair3       & in           & -0.18      & ***\\
      Qwen2   & pair3       & people       & -0.21      & ***\\
      Qwen2   & pair4       & in           & -0.09      & **\\
      Qwen2   & pair4       & people       & -0.05      & \\
      Qwen2   & pair5       & in           & -0.18      & ***\\
      Qwen2   & pair5       & people       & -0.36      & ***\\
      \bottomrule
 \hline
\end{tabular}
\caption{Correlation results for the WVS dataset using the Qwen2-0.5B model: analysis reveals significant negative correlations across all token pairs, with the most pronounced being -0.36 for pair5. The presence of a single positive correlation at 0.14*** for pair2 in the 'in' mode provides a contrast to the generally negative trend.}
    \label{tab:table7} 
\end{table}

\begin{table}[ht] 
\begin{tabular}{ l l l S l }  
\toprule
\emph{Model} & \emph{Tokens} & \emph{Mode}  & \emph{r} & \emph{p-value} \\
\midrule
      Qwen2   & pair1       & in           &  0.11     & *\\
      Qwen2   & pair1       & people       &  0.11     & *\\
      Qwen2   & pair2       & in           & -0.06     & \\
      Qwen2   & pair2       & people       & -0.26     & ***\\
      Qwen2   & pair3       & in           &  0.30     & ***\\
      Qwen2   & pair3       & people       &  0.14     & **\\
      Qwen2   & pair4       & in           & -0.18     & **\\
      Qwen2   & pair4       & people       & -0.22     & ***\\
      Qwen2   & pair5       & in           & -0.38     & ***\\
      Qwen2   & pair5       & people       & -0.35     & ***\\
      \bottomrule
 \hline
\end{tabular}
\caption{Correlation results for the PEW dataset using the Qwen2-0.5B model: analysis demonstrates a mix of positive and negative correlations. Highlights include a strong positive correlation of 0.30*** for pair3, contrasting with significant negative correlations, especially for pair5 under both prompt with each reaching beyond -0.35.}
    \label{tab:table8} 
\end{table}

The results from the Qwen2-0.5B model, as detailed in Tables~\ref{tab:table7} and~\ref{tab:table8}, are less favorable than those obtained with the OPT model, presenting weaker correlations between the model outputs and the survey scores. These results predominantly show statistically significant and largely negative correlations across the different token pairs, particularly within the WVS dataset. There appears to be a consistent pattern where the choice of moral token generally has a more substantial impact on the score than the prompt mode used. Notably, the highest moral scores recorded are 0.14 in the WVS dataset and 0.30 in the PEW dataset, both achieving a 99.9{\%} significance level.

The BLOOMZ-560M model has produced the best results so far in terms of alignment between the model outputs and the survey scores. Using the WVS questions as prompts, the average moral scores are 0.25*** and 0.29***, both significant and the highest recorded thus far across the averaged token pairs scores. Similarly, when using the topics and countries from PEW, the scores are 0.16** and 0.11* for the two prompt types.

\begin{table}[!htbp] 
\begin{tabular}{ l l l S l }  
\toprule
\emph{Model} & \emph{Tokens} & \emph{Mode}  & \emph{r} & \emph{p-value} \\
\midrule
      BLOOM   & pair1       & in           & -0.07     & **\\
      BLOOM   & pair1       & people       & -0.16     & ***\\
      BLOOM   & pair2       & in           &  0.14     & ***\\
      BLOOM   & pair2       & people       &  0.12     & ***\\
      BLOOM   & pair3       & in           & -0.04     & \\
      BLOOM   & pair3       & people       & -0.36     & ***\\
      BLOOM   & pair4       & in           &  0.36     & ***\\
      BLOOM   & pair4       & people       &  0.26     & ***\\
      BLOOM   & pair5       & in           &  0.21     & ***\\
      BLOOM   & pair5       & people       &  0.30     & ***\\
      \bottomrule
 \hline
\end{tabular}
\caption{Correlation results for the WVS dataset using the BLOOMZ-560M model: analysis showcases a predominance of significant, strong positive correlations, with three token pairs for both prompts achieving these results, with the highest reaching 0.36***. Negative correlations, while present, are less pronounced.}
    \label{tab:table9} 
\end{table}

\begin{table}[ht] 
\begin{tabular}{ l l l S l }  
\toprule
\emph{Model} & \emph{Tokens} & \emph{Mode}  & \emph{r} & \emph{p-value} \\
\midrule
      BLOOM   & pair1       & in           & -0.25     & ***\\
      BLOOM   & pair1       & people       & -0.13     & *\\
      BLOOM   & pair2       & in           &  0.08     & \\
      BLOOM   & pair2       & people       &  0.12     &  *\\
      BLOOM   & pair3       & in           & -0.07     & \\
      BLOOM   & pair3       & people       &  0.12     & *\\
      BLOOM   & pair4       & in           &  0.28     & ***\\
      BLOOM   & pair4       & people       &  0.23     & ***\\
      BLOOM   & pair5       & in           &  0.16     & **\\
      BLOOM   & pair5       & people       &  0.08     & \\
      \bottomrule
 \hline
\end{tabular}
\caption{Correlation results for the PEW dataset using the BLOOMZ-560M model: analysis highlights a predominance of strong positive correlations, more pronounced than those seen with earlier models. Significant positive results include correlations of 0.28 and 0.23 both for pair4. While negative correlations are present, they are comparatively milder.}
    \label{tab:table10} 
\end{table}

As illustrated in Table~\ref{tab:table9}, the results for the WVS dataset showcase a prevalence of significant, strong positive correlations, surpassing the performance of previous models. Three token pairs achieve these notable results for both prompts, with the highest recorded at 0.36***. Although negative correlations are present, they are comparatively less pronounced. 

Similarly, for the PEW dataset, the results are also encouraging as shown in Table~\ref{tab:table10}. Significant positive correlations prevail, though they are not as high as those observed for the WVS dataset. The highest positive correlation recorded is 0.28***, while the negative correlations, that are present in the dataset, lack statistical significance.

\subsection{Distribution of moral scores per topic}

As depicted in Figure~\ref{fig2} in section 3, the spread of responses varies significantly across different moral topics. Topics such as 'for a man to beat his wife', 'stealing property', and 'violence against other people' show limited variation across countries and are mainly positioned on the left side, indicating negative moral scores. In contrast, topics like 'homosexuality', 'sex before marriage', and 'having casual sex' exhibit a wide range of responses, spanning both negative and positive moral scores.

\begin{figure}[ht]
\centering
\includegraphics[width=0.4\textwidth]{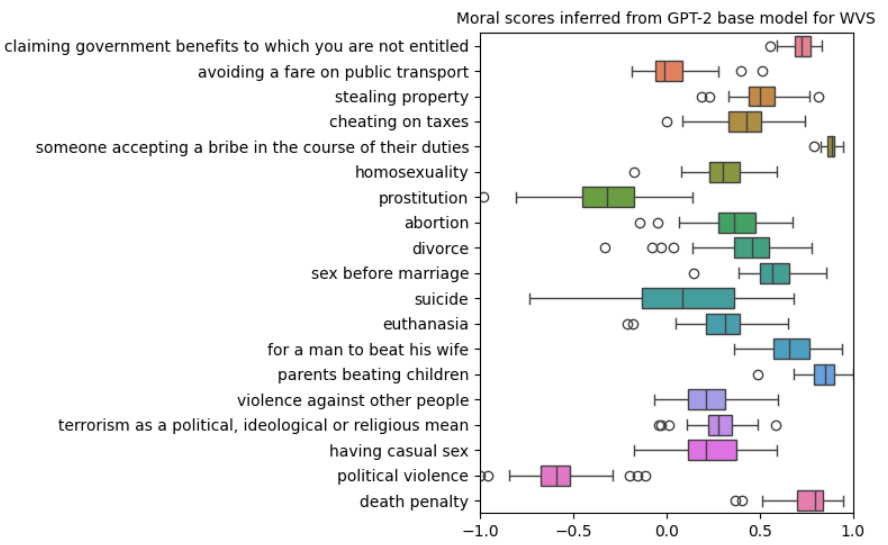}
\caption{Distribution of normalized moral scores from GPT-2 base model using the WVS dataset}
\label{fig5}
\end{figure}

When comparing the moral scores from the GPT-2 base model for the WVS dataset, as depicted in Figure~\ref{fig5}, with the actual survey responses, notable differences emerge. The GPT-2-derived scores predominantly gather on the right side of the x-axis, indicating a tendency toward positive moral judgments. Topics such as 'sex before marriage', 'having casual sex', 'homosexuality', and 'divorce' exhibit wide variation according to the survey, highlighting diverse viewpoints among people from different countries. In contrast, the GPT-2 model shows less variation for most topics, but notable disagreements are seen in topics 'suicide' and 'prostitution'.

Interestingly, despite the general trend of smaller variations in GPT-2 scores—which aligns with findings from previous studies—the model also unexpectedly shows a significant number of positive moral judgments across different prompts and token pairs. This suggests that while the model captures some aspects of human moral reasoning, its application still presents challenges in accurately mirroring the complex landscape of human moral values.

\begin{figure}[ht]
\centering
\includegraphics[width=0.4\textwidth]{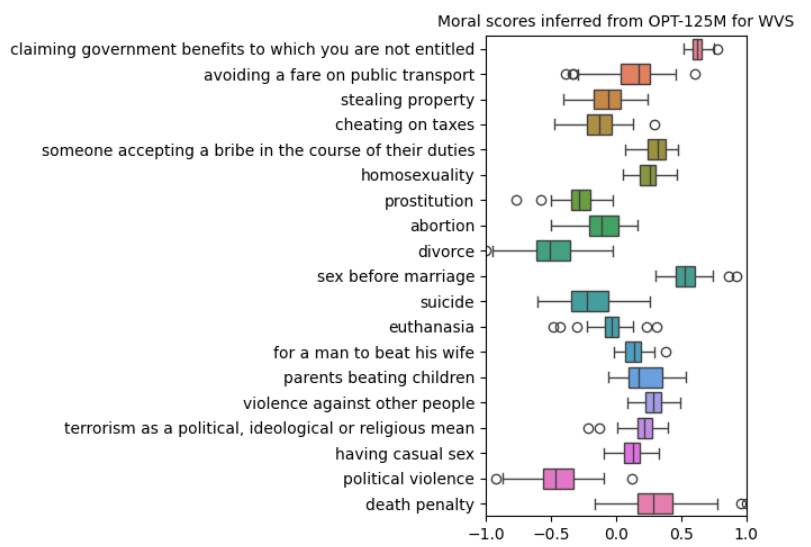}
\caption{Distribution of normalized moral scores from OPT-125M model using the WVS dataset}
\label{fig6}
\end{figure}

The moral scores inferred from the OPT-125 model for the WVS dataset reveal that for certain topics, the results closely follow those of the GPT-2 base model, as shown in Figure~\ref{fig6}.  However, for topics where the behavior diverges from that observed in the GPT-2 model, the scores from the OPT-125 model tend to cluster closer to zero rather than extending into more positive values. This suggests a more neutral stance by the OPT-125 model on these particular issues.

\begin{figure}[ht]
\centering
\includegraphics[width=0.4\textwidth]{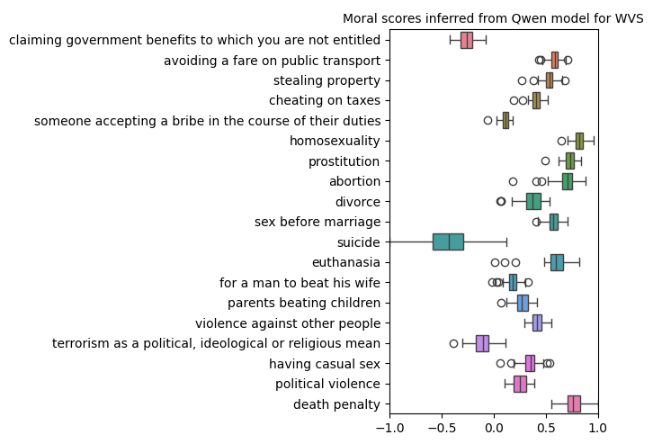}
\caption{Distribution of normalized moral scores from Qwen2 model using the WVS dataset}
\label{fig7}
\end{figure}

The results from the Qwen2 model (Figure~\ref{fig7}), when assessed using WVS moral topics as prompts, show patterns similar to those observed with the OPT model. Variations in moral scores are much smaller than those seen in the survey data. Additionally, the boxplots are predominantly positioned on the positive side of the x-axis, indicating a bias towards viewing these actions as morally acceptable, which does not align with the societal views as they are reflected in the survey results.

\begin{figure}[ht]
\centering
\includegraphics[width=0.4\textwidth]{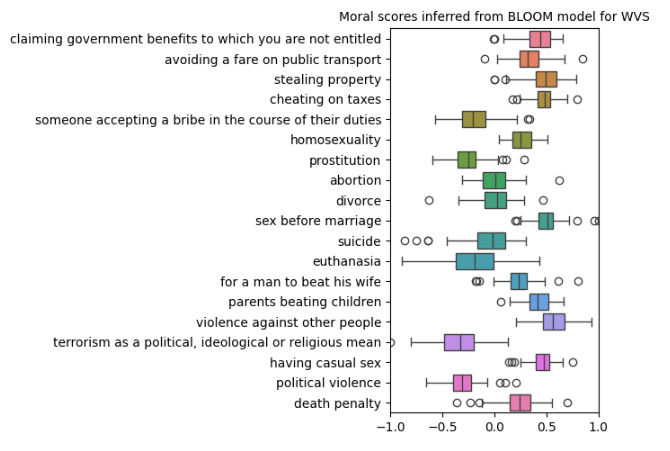}
\caption{Distribution of normalized moral scores from BLOOMZ-560M model using the WVS dataset}
\label{fig8}
\end{figure}

As highlighted by the significant positive correlations between the BLOOM model's scores and the survey results, as described earlier, the BLOOM model performs better in mirroring societal views. As displayed in Figure~\ref{fig8}, it exhibits greater variability in moral scores compared to previous models, and these scores are now more closely aligned with the actual survey responses, tending towards more negative assessments. This shift suggests that this model offers a more accurate representation of societal views as presented in the survey data.

Regarding the PEW survey, as depicted in Figure~\ref{fig4} in section 3, responses to moral questions display significant diversity, similar to those in the WVS survey. Notably, topics such as 'married people having an affair' and 'gambling' consistently receive negative judgments, while they also show significant disagreement among respondents from different countries. 'Homosexuality' and 'sex between unmarried adults' exhibit the greatest variability, underscoring sharp differences in moral views across populations.

\begin{figure}[ht]
\centering
\includegraphics[width=0.4\textwidth]{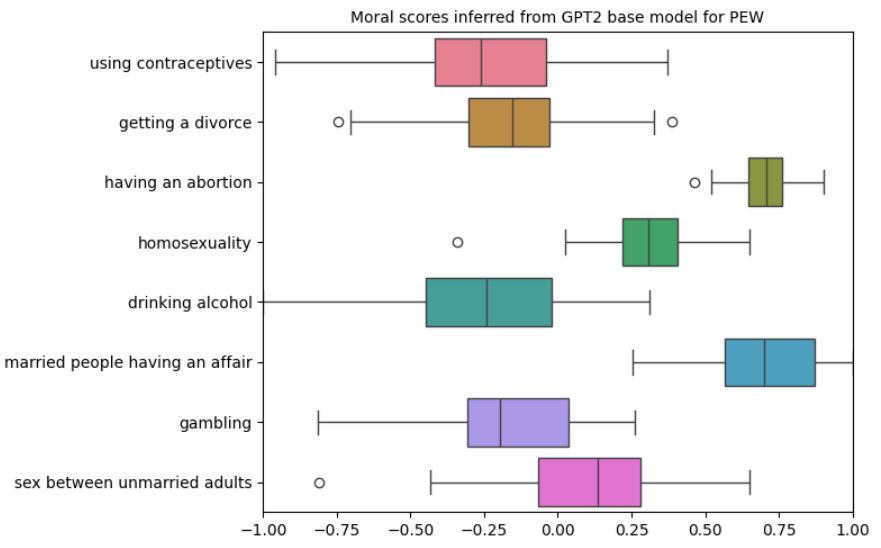}
\caption{Distribution of normalized moral scores from GPT-2 base model using the PEW dataset}
\label{fig9}
\end{figure}

When comparing the PEW survey results to the outputs from the GPT-2 base model for the same dataset, as shown in Figure~\ref{fig9}, several striking differences emerge. Firstly, the model demonstrates unexpectedly high variations in its responses, which contrasts with similar studies. There is a clear disagreement on topics such as 'married people having an affair' and 'getting an abortion'; the model typically sees these actions as acceptable, whereas the people who participated in the survey categorize them as mostly unacceptable. The only topic for which the model and the survey agree is 'drinking alcohol,' since both exhibit significant variability.

\begin{figure}[ht]
\centering
\includegraphics[width=0.4\textwidth]{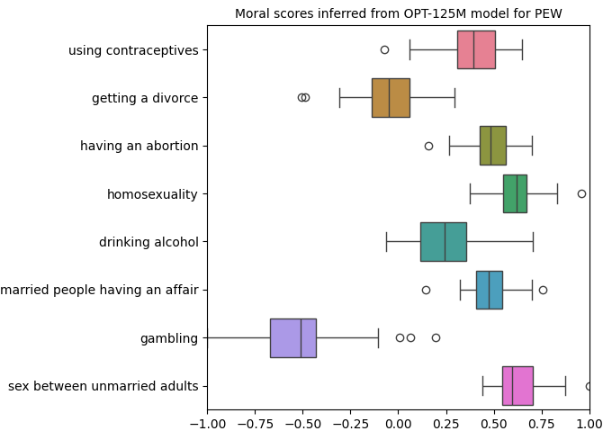}
\caption{Distribution of normalized moral scores from OPT-125M model using the PEW dataset}
\label{fig10}
\end{figure}

The OPT-125M model's moral scores for the PEW dataset, as illustrated in  Figure~\ref{fig10}, generally show smaller variations across most topics compared to the GPT-2 base model, with the notable exception of 'gambling,' which displays significant spread. Additionally, like the GPT-2, the scores are predominantly shifted towards the positive side, suggesting a more favorable moral assessment of most topics.

\begin{figure}[ht]
\centering
\includegraphics[width=0.4\textwidth]{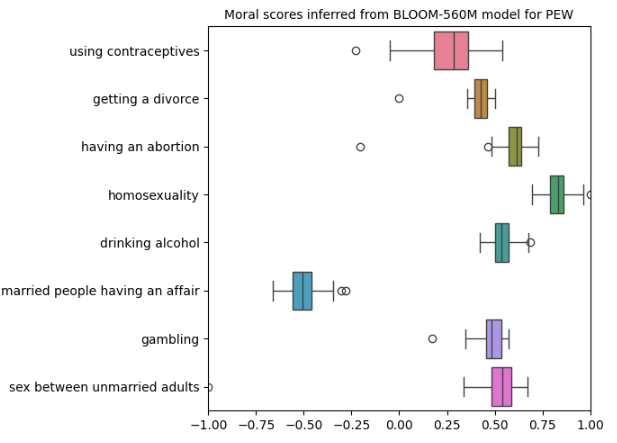}
\caption{Distribution of normalized moral scores from BLOOMZ-560M model using the PEW dataset}
\label{fig11}
\end{figure}

For the BLOOMZ-560M model (Figure~\ref{fig11}), variations in moral scores are even smaller across most topics, with only a few outliers. Notably, the topic 'married people having an affair' is consistently considered unjustifiable in all countries, according to the model's assessments.

\begin{figure}[ht]
\centering
\includegraphics[width=0.4\textwidth]{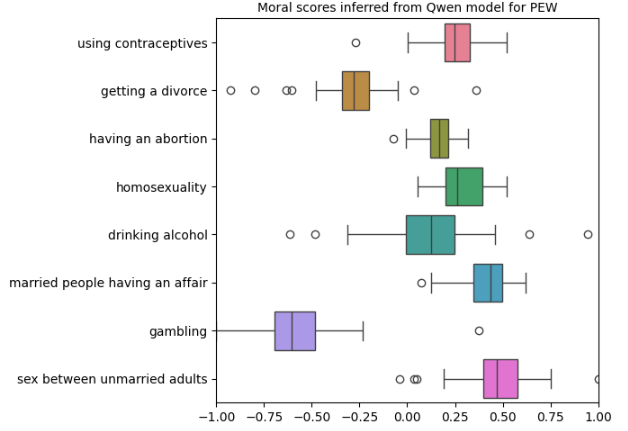}
\caption{Distribution of normalized moral scores from Qwen2 model using the PEW dataset}
\label{fig12}
\end{figure}

Results from the Qwen2 model are similar to those from the BLOOM model but are slightly shifted closer to zero, indicating a more neutral stance on the issues (Figure~\ref{fig12}).

\section{Conclusion and Discussion}

The aim of this study was to investigate whether pre-trained monolingual and multilingual language models contain knowledge about moral norms across many different cultures. The analysis shows that the examined LLMs do capture certain cultural value differences, but these only weakly align with established values surveys. They tend to characterize most topics as justifiable or generally acceptable across most countries, which contrasts with the varied and often contradictory views reflected in the WVS and the PEW survey data. While LLMs are capable of processing language, they cannot fully perceive the complex societal and cultural contexts that influence moral judgments.

The outputs of the four models reveal notable differences in the variability and alignment of moral scores compared to the actual survey results. The correlation scores between the models and the survey data were largely disappointing, with predominantly negative values, most of which were statistically significant. Furthermore, an in-depth analysis conducted by calculating correlations separately for the different prompt types and token pairs did not provide solid conclusions, as the results exhibited substantial variability across models and datasets.

The variant of BLOOM, BLOOMZ-560M showed a closer approximation to human judgments by aligning more consistently with negative assessments than the other three models. Yet it still failed to reflect human opinions even to a moderate degree. A possible reason for this performance could be attributed to its multilingual capabilities. As a multilingual model, BLOOM is trained on diverse linguistic datasets, which potentially enables it to access more cultural and moral contexts compared to monolingual models. Additionally, the performance of BLOOM, which is similar to that of GPT-3—a significant improvement over GPT-2—has been trained on 46 different languages and 13 programming languages in total. In contrast, Qwen2, the other multilingual model used in this study, did not showcase similar performance. This may be due to it being trained on data in fewer languages, with a particular focus on Chinese and English.

Another conclusion is that the four LLMs tend to characterize most topics as generally acceptable. Language models may simplify complex moral judgments due to their inability to fully understand nuanced cultural contexts and ethical considerations. As a result, they tend to adopt a more generalizable and justifiable stance. Additionally, without specific context, the models might be designed to lean towards more neutral or positive judgments to avoid controversial or negative outputs, which could be seen as safer or more acceptable. Thus, to avoid drawing solid conclusions, it's important to recognize that the models' responses might not accurately represent reality due to their lack of sufficient contextual information.

As a final conclusion, from the different experiments with the prompt modes and the different token pairs it was concluded that the choice of moral tokens used has a greater impact on the model scores than the choice of prompt types. This indicates that the selection of moral tokens substantially influences how the models assess moral norms.

Furthermore, it is worth noting that using alternative correlation coefficient metrics, such as Spearman's rank correlation coefficient, and testing newer models like GPT-3 or GPT-4, could potentially lead to different conclusions. Even deploying the available versions of the four models used in this study with the highest number of parameters could give different results. These methodological and architectural variations might offer additional insights into how language models interpret and generate moral judgments. Further exploration following these adjustments is essential to improve our understanding of their capabilities and limitations in ethical reasoning and comprehension.

\section{Limitations}

Although the datasets employed are publicly available and include responses from participants across different countries, they cannot fully represent the moral norms of all cultural groups globally or predict how these norms might evolve over time (Bloom, 2010; Bicchieri, 2005). Moreover, this study only explores a limited range of moral issues per country, and thus should not be considered exhaustive of the moral dilemmas people face worldwide. Additionally, averaging moral ratings for each culture simplifies the diverse range of moral values to a single value, which is a limitation of this study.

Furthermore, computational limitations constrained the scope of this research. The computational demands of the models were significant, and the availability of tools offering free additional resources restricted the analysis. Similarly, the use of more advanced models like GPT-3 or GPT-4 was not possible due to their requirement for paid access. Consequently, this restriction likely impacted the comprehensiveness of the findings and the depth of the analysis.

\bibliographystyle{acl_natbib}
\bibliography{acl2023}

\section*{Acknowledgements}
I would like to express my gratitude to my supervisors, Professor Ayoub Bagheri and Hadi Mohammadi, for their guidance, support, and encouragement throughout this research. Their insights and constructive feedback were crucial to the successful completion of this research.

I also acknowledge the authors of the papers I tried to replicate, as well as the creators of the datasets and tools utilized in this research, including the World Values Survey and the PEW Research Center.

\appendix

\section{Appendix}
\label{sec:appendix}

\begin{table}[!htbp]
\begin{tabular}{ l l l S l }  
\toprule
\emph{Model} & \emph{Tokens} & \emph{Mode}  & \emph{r} & \emph{p-value} \\
\midrule
      GPT2-M   & pair1       & in           & -0.35     & *** \\
      GPT2-M   & pair1       & people       & -0.04     & *** \\
      GPT2-M   & pair2       & in           &  0.01     & *\\
      GPT2-M   & pair2       & people       &  0.16     & \\
      GPT2-M   & pair3       & in           & -0.18     & *** \\
      GPT2-M   & pair3       & people       & -0.18     & *** \\
      GPT2-M   & pair4       & in           &  0.11     & \\
      GPT2-M   & pair4       & people       & -0.17     & \\
      GPT2-M   & pair5       & in           & -0.04     & \\
      GPT2-M   & pair5       & people       & -0.33     & **\\
      \bottomrule
 \hline
\end{tabular}
\caption{Correlation results for the WVS dataset using the GPT-2-Medium model: analysis shows almost exclusively negative correlations, with half of these being statistically significant. This table demonstrates a higher incidence of negative scores compared to those observed using the GPT-2 base model. The results are quite similar to those obtained by the GPT-2-Large model for the same dataset.}
    \label{tab:table11} 
\end{table}

\begin{table}[htbp]
\begin{tabular}{ l l l S l }  
\toprule
\emph{Model} & \emph{Tokens} & \emph{Mode}  & \emph{r} & \emph{p-value} \\
\midrule
      GPT2-M   & pair1       & in           & -0.25     & *** \\
      GPT2-M   & pair1       & people       &  0.11     & *** \\
      GPT2-M   & pair2       & in           &  0.12     & *\\
      GPT2-M   & pair2       & people       & -0.01     & \\
      GPT2-M   & pair3       & in           &  0.26     & *** \\
      GPT2-M   & pair3       & people       &  0.35     & *** \\
      GPT2-M   & pair4       & in           &  0.19     & \\
      GPT2-M   & pair4       & people       & -0.04     & \\
      GPT2-M   & pair5       & in           &  0.04     & \\
      GPT2-M   & pair5       & people       & -0.19     & ***\\
      \bottomrule
 \hline
\end{tabular}
\caption{Correlation results for the PEW dataset using the GPT-2-Medium model: analysis indicates a range of correlations from positive to negative. Findings include a strong positive correlation for pair3 and significant negative correlations for pair1 and pair5. These results suggest varying alignment between the model scores and survey responses across different contexts.}
    \label{tab:table12} 
\end{table}

\end{document}